\documentclass[letterpaper, 10 pt, conference]{ieeeconf}  

\IEEEoverridecommandlockouts                              

\usepackage{cite}
\usepackage{multicol}
\usepackage{import}
\usepackage{soul}
\usepackage{xcolor}
\usepackage{wrapfig}
\usepackage{xspace}
\usepackage{graphicx}
\usepackage{multirow}
\usepackage{rotating}
\usepackage{amssymb}
\usepackage{algorithm}
\usepackage{multirow}
\usepackage{array}
\usepackage{makecell}
\usepackage{svg}

\usepackage[noend]{algpseudocode}
\usepackage{amsmath}
\usepackage{booktabs}
\usepackage{siunitx}
\newcommand{\open}{\ensuremath{\textit{OPEN}}\xspace}
\newcommand{\closed}{\ensuremath{\textit{CLOSED}}\xspace}

\newcommand{\state}{\ensuremath{\mathit{s}}\xspace}

\newcommand{\edge}{\ensuremath{\mathit{e}}\xspace}

\newcommand{\pachs}{\textsc{Pachs}}

\newcommand{\tps}{\ensuremath{\textit{Panda-Shelf}}\xspace}
\newcommand{\ttf}{\ensuremath{\textit{PushT-Fixed}}\xspace}
\newcommand{\ttr}{\ensuremath{\textit{PushT-Rand}}\xspace}
\newcommand{\tto}{\ensuremath{\textit{PushT-Obs}}\xspace}

\newcommand{\rollout}{\ensuremath{\textit{Single Rollout}}\xspace}
\newcommand{\prollout}{\ensuremath{\textit{Parallel Rollout}}\xspace}
\newcommand{\pbeam}{\ensuremath{\textit{Parallel Beam Search}}\xspace}
\newcommand{\epase}{\ensuremath{\textit{ePA*SE}}\xspace}

\usepackage{hyperref}

\title{\LARGE \bf
Parallel Heuristic Search as Inference \\ for Actor-Critic Reinforcement Learning Models}

\author{\authorblockN{\textbf{Hanlan Yang$^{*1,2}$, Itamar Mishani$^{*1}$, Luca Pivetti$^{1,3}$, Zachary Kingston$^{2}$, and Maxim Likhachev$^{1}$}}
\authorblockA{$^1$Carnegie Mellon University \qquad 
$^2$Purdue University \qquad
$^3$University of Milano-Bicocca \\
$*$ Equal Contribution
}}

\begin{document}

\maketitle
\thispagestyle{empty}
\pagestyle{empty}

\begin{abstract}

Actor-Critic models are a class of model-free deep reinforcement learning (RL) algorithms that have demonstrated effectiveness across various robot learning tasks. While considerable research has focused on improving training stability and data sampling efficiency, most deployment strategies have remained relatively simplistic, typically relying on direct actor policy rollouts. In contrast, we propose \pachs{} (\textit{P}arallel \textit{A}ctor-\textit{C}ritic \textit{H}euristic \textit{S}earch), an efficient parallel best-first search algorithm for inference that leverages both components of the actor-critic architecture: the actor network generates actions, while the critic network provides cost-to-go estimates to guide the search. Two levels of parallelism are employed within the search---actions and cost-to-go estimates are generated in batches by the actor and critic networks respectively, and graph expansion is distributed across multiple threads. We demonstrate the effectiveness of our approach in robotic manipulation tasks, including collision-free motion planning and contact-rich interactions such as non-prehensile pushing. Visit \href{https://p-achs.github.io}{p-achs.github.io} for demonstrations and examples.

\end{abstract}

\section{Introduction}


Reinforcement Learning (RL) has become a central paradigm for robot control, offering a way to learn behaviors that are difficult to manually specify through cost functions, dynamics models, or action abstractions. Despite this strength, RL methods still face significant challenges in generalization during inference and in solving complex problems that are common in robotic manipulation. Much of the prior work has focused on improving model architectures and training strategies, but comparatively little attention has been given to inference strategies. As a result, RL models are typically deployed as one-step predictors during execution, lacking the ability to perform multi-step forward reasoning or backtracking.

\begin{figure}[h] 
    \centering
    \includegraphics[width=0.65\linewidth]{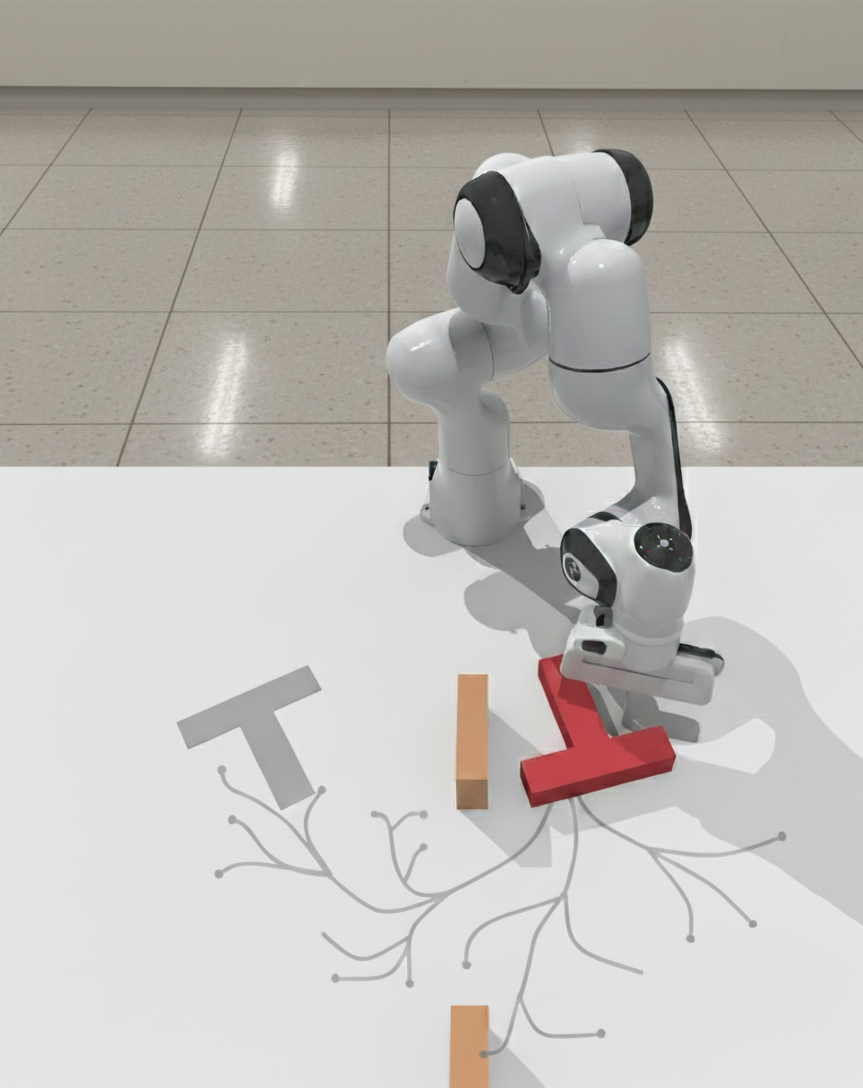}
    \caption{\pachs{} leverages both actor and critic networks in a parallel best-first search: the actor generates candidate actions while the critic provides learned heuristics to guide exploration through the state space. Here shown for the push-T task, our algorithm builds an implicit lattice graph to find the trajectory for the robotic arm to manipulate a T-shaped object to a target pose.}
    \label{fig:pusht}
    \vspace{-2em}
\end{figure}

Consider the example shown in Fig.~\ref{fig:pusht}, where the objective is to push the T-shaped object to a target pose. While specifying and modeling this task with classical model-based planning algorithms is challenging due to complex contact dynamics, an RL model can learn non-prehensile manipulation behaviors through a simple reward function.
Such models perform well in relatively simple settings, for example, when the goal is fixed and the environment is uncluttered. However, as problems become more complex and require inference generality (e.g., when multiple objects must be manipulated sequentially, or when goals depend on environmental constraints), RL often fails to learn sufficiently robust policies. In these cases, search methods provide an alternative: they can explicitly plan multiple steps into the future, enabling exploration of the state space through branching and backtracking.

The main goal of this work is to combine the advantages of traditional planning algorithms with those of learned models.
Specifically, this work integrates a Soft Actor–Critic (SAC) model, trained to control a robotic arm, within a best-first heuristic search framework. In this integration, the actor network proposes candidate actions (i.e., edges) while the critic network serves as a learned heuristic function.  
The reinforcement learning approach underlying SAC enables the model to capture complex information that is often difficult to represent with hand-crafted action spaces, heuristics and cost functions, such as robot-object interactions, dynamics, and environmental constraints. 
Furthermore, to improve efficiency, we parallelize the exploration phase of the search process by assigning separate threads to each expansion step, and, if possible, by evaluating edges in parallel. 

We evaluate the proposed approach on two different applications. The first scenario involves a shelf environment where the robotic arm must move its end effector to a desired spatial position along a collision-free path. 
The second scenario presents a push-T task, where the arm must interact with a T-shaped object to push it to a specific position and orientation. 
Here, we trained the model in an obstacle-free environment and then applied the algorithm in environments with added obstacles to study generalizability. The results demonstrate substantial performance improvements, enabling the effective use of imperfectly trained models in complex scenarios through search--a common challenge in RL deployment. 

Our contributions are:
\begin{itemize}
    \item A novel algorithm, \pachs{}, enabling best-first search planning with learned actor-critic RL models.
    \item Multi-layered parallelization strategies (CPU thread-level and GPU batch-level) that achieve significant computational efficiency gains.
    \item Experimental evaluations, demonstrating how \pachs{} improves the deployment and generalization of RL models by enabling robust performance in complex environments.
\end{itemize}
\section{Related Work}



Our approach combines reinforcement learning (RL) with heuristic search algorithms, building on extensive prior work in both domains. Actor-Critic methods like SAC have proven effective for learning complex robotic behaviors through the combination of policy-based and value-based techniques~\cite{haarnoja2018soft}. However, these methods typically lack the structural exploration capabilities of search algorithms when deployed. Conversely, classical search methods like A* \cite{astar} provide robust planning guarantees but struggle with continuous action spaces and complex robotic dynamics that are difficult to capture with hand-crafted heuristics.
Using search to improve zero-shot generalization in reinforcement learning is not the only approach (see~\cite{kirk2023survey} for an overview); however, compared to other approaches, this paper proposes an algorithm that does not require retraining and can be used with no modification to the policy.

\subsection{Combining Learning and Search}

The integration of learning and search has been most prominently demonstrated in domains where simulating rollouts is trivial, such as in game-playing domains. AlphaGo~\cite{silver2016mastering} combines a policy network, value network, and Monte Carlo Tree Search (MCTS) in an iterative process: the policy network proposes promising moves, the value network estimates expected outcomes of future board states, and MCTS performs lookahead search to refine decision-making. While conceptually similar to our approach, several key differences distinguish our work: AlphaGo operates in discrete environments, significantly reducing the complexity of state expansion and search; computations are processed sequentially, making each decision computationally expensive; and the policy and value networks are trained independently, requiring separate training procedures.

Learning heuristic functions has been explored extensively, particularly for domains where designing effective heuristics is challenging. For instance, DeepCubeA~\cite{agostinelli2019solving} uses a deep neural network to learn heuristic functions for solving Rubik's Cube, which are then integrated into an A*-like search algorithm. Similar approaches have been applied to various combinatorial problems~\cite{arfaee2011learning}, but these methods are typically restricted to value-based approaches and discrete action spaces. Recent work has developed new priority functions for search that integrate learned components, such as local heuristics~\cite{local_heuristic}. Other, learned heuristics expansion method such as \cite{bokan2024slope}, focus on learning to guide node expansion, using learned models to prune unfavorable nodes during search.

More recently, several works have explored combining policy-based methods with search. AlphaZero~\cite{silver2017mastering} extends the AlphaGo framework to learn both policy and value functions from scratch through self-play. MuZero~\cite{schrittwieser2020mastering} further advances this by learning a model of the environment dynamics. However, these approaches remain focused on discrete domains and sequential computation, limiting their applicability to continuous robotic control problems where parallel execution and real-time constraints are critical.

Our work addresses these limitations by integrating SAC, which naturally handles continuous action spaces, with parallel A*-like search. Unlike prior approaches that require independent training of policy and value networks, we leverage the joint training inherent in actor-critic methods. Furthermore, our parallel search framework enables efficient exploration in continuous domains, making the approach practical for real-time robotic applications.

\subsection{Heuristic Search and Parallelization}

Advances in computing hardware, such as the increasing number of cores in modern processors, have driven significant efforts to parallelize best-first search algorithms, which inherently require extensive exploration through node expansion and edge evaluation. 
In robotic domains, edge evaluations are particularly expensive, involving collision checking, physics simulation, and complex dynamics computations, making parallelization essential for practical performance \cite{shohin_thesis}\cite{vamp}.

Search-based methods, like A* and Beam search, can leverage parallelization by generating successors concurrently during state expansion, though their scalability is often constrained by the domain’s branching factor. For instance, Beam search~\cite{beamsearch}, which traverses a graph layer by layer, selects a fixed number of nodes (the beam width) at each layer for expansion. This expansion work can be distributed across multiple workers. However, parallel Beam search remains inherently synchronous because selecting nodes for expansion requires constructing the complete candidate list beforehand. Monte Carlo Tree Search (MCTS) has also been explored for parallelization. PMBS~\cite{huang2022parallel}, for example, parallelizes the simulation rollout process in batches. However, MCTS requires multiple cycles of simulation roll-out and backpropagation to update the heuristic for a single-step decision-making, which makes it computationally intensive and limits the planning horizon given the same computation budget when compared to other search methods.



Heuristic best-first search algorithms, particularly A* and its variants, are non-trivial to parallelize due to their sequential expansion nature and the need to maintain theoretical guarantees. Early approaches like Parallel A*~\cite{irani1986parallel} use a centralized \open list (priority queue) and assign the current best states to available CPU cores. PRA*~\cite{bonet2001planning} and HDA*~\cite{kishimoto2009scalable} improve upon this by giving each CPU core its own \open list, with HDA* adding an asynchronous message passing system to reduce communication blocking.
GPU implementations follow similar principles, maintaining multiple parallel \open lists~\cite{he2021efficient, zhou2015massively}. However, all these approaches must allow states to be re-expanded to guarantee optimality, and the number of re-expansions can grow exponentially with parallelization degree, especially with weighted heuristics, causing significant performance degradation~\cite{mukherjee2022epa, phillips2014pa}.

More recent work addresses these limitations through alternative parallelization strategies. PA*SE~\cite{phillips2014pa} waives the re-expansion requirement by estimating state independence using admissible pairwise heuristics. The ePA*SE family~\cite{mukherjee2023gepa, mukherjee2022epa, yang2023epa} further improves efficiency by decoupling state expansion from edge evaluation, particularly beneficial for robotic domains with expensive edge evaluations (collision checking, simulation queries). However, these methods still face challenges in continuous robotic domains where physics simulation and contact-rich interactions cannot be easily delegated or approximated.
While parallelization advances improve computational efficiency, they still rely on hand-crafted heuristics, action spaces, and cost functions that may not capture the complex dynamics of manipulation tasks. 

\section{Preliminaries}

This section establishes the theoretical foundation for integrating reinforcement learning with heuristic search. We begin by formulating continuous robotic control as a Markov Decision Process (MDP), then review Actor-Critic methods and heuristic search algorithms. Finally, we demonstrate the fundamental connection between these paradigms that enables our integration approach.

\subsection{Problem Formulation} \label{sec:prob_form}

We consider robotic manipulation tasks modeled as Markov Decision Processes $(\mathcal{S}, \mathcal{A}, f, R, \gamma)$, where $\mathcal{S} \subset \mathbb{R}^d$ and $\mathcal{A} \subset \mathbb{R}^m$ represent continuous state and action spaces, $f: \mathcal{S} \times \mathcal{A} \rightarrow \mathcal{S}$ defines the system dynamics, $R: \mathcal{S} \times \mathcal{A} \rightarrow \mathbb{R}$ is the reward function, and $\gamma \in [0, 1]$ is the discount factor. Given a trained Soft Actor-Critic model consisting of an actor network $\pi_{\theta}: \mathcal{S} \rightarrow \mathcal{P}(\mathcal{A})$ (where $\mathcal{P}(\mathcal{A})$ denotes probability distributions over actions) and critic network $Q_{\phi}: \mathcal{S} \times \mathcal{A} \rightarrow \mathbb{R}$, and given an initial state $s_0 \in \mathcal{S}$ and goal region $\mathcal{G} \subseteq \mathcal{S}$, we seek to find a feasible trajectory $\tau = (s_0, a_0, s_1, a_1, \ldots, s_T)$ that reaches a goal state while minimizing accumulated cost. Formally, we aim to solve:
\begin{align}
    \min_{\tau} \quad &\sum_{t=0}^{T-1} c(s_t, a_t) \label{eq:cost-objective}\\
    \text{s.t.} \quad &s_T \in \mathcal{G} \label{eq:goal-constraint}\\
    &s_t \in \mathcal{S}_{\text{valid}} \quad \forall t \in \{0, 1, \ldots, T\} \label{eq:state-constraint}\\
    &s_{t+1} = f(s_t, a_t) \quad \forall t \in \{0, 1, \ldots, T-1\} \label{eq:dynamics-constraint}
\end{align}
where $c(s,a) = -R(s,a)$ converts rewards to costs and $\mathcal{S}_{\text{valid}} \subseteq \mathcal{S}$ denotes the constraint-satisfying state space (e.g., collision-free configurations).



To solve this optimization problem, we leverage reinforcement learning models that have learned both action selection strategies and value estimation from interaction data. Specifically, we build upon actor-critic architectures that provide both components needed for effective search: action generation and state evaluation.

\subsection{Actor-Critic RL}

Actor-critic methods~\cite{konda1999actor,sutton2018reinforcement} address the limitations of pure policy gradient and value-based approaches by maintaining two complementary function approximators: an actor network $\pi_{\theta}$ that parameterizes the policy, and a critic network $Q_{\phi}$ that estimates the state-action value (Q-) function
\begin{equation}
    Q^{\pi}(s, a) = \mathbb{E}_{\tau \sim \pi}\left[\sum_{t=0}^{\infty} \gamma^t R(s_t, a_t) \mid s_0 = s, a_0 = a\right], \nonumber
\end{equation}
representing the expected cumulative reward when taking action $a$ in state $s$ and following policy $\pi$ thereafter. This dual architecture enables stable learning through policy iteration, where the critic evaluates the current policy, and the actor improves the policy using gradients derived from the critic's evaluations.


Standard deployment of actor-critic models utilizes only the actor network for action selection, effectively discarding the critic's learned value assessments once training concludes \cite{haarnoja2018soft}. This represents a significant underutilization of the available learned knowledge, as the critic contains valuable information about state quality and expected future returns that could inform more sophisticated decision-making during execution.

To fully utilize both components of the actor-critic model, we integrate them within a systematic search framework. Best-first search algorithms provide the structured exploration needed to solve the optimization problem while allowing us to leverage learned knowledge from both networks.

\subsection{Best-first Search}

Best-first search algorithms, particularly A* and its variants~\cite{hart1968formal, aine2016multi, mukherjee2022epa}, provide algorithmic approaches to solving the optimization problem defined in Equations~\eqref{eq:cost-objective}-\eqref{eq:dynamics-constraint}. These algorithms construct an implicit graph $G = (V, E)$ where vertices $v \in V$ represent states $s \in \mathcal{S}$ and edges $e \in E$ represent state-action pairs $(s, a)$ that induce transitions $s' = f(s, a)$.

The A* algorithm maintains a priority queue of discovered states, ordered by an evaluation function:
\begin{equation} 
    \label{eq:f_value}
    f(s) = g(s) + h(s), \nonumber
\end{equation}
where $g(s)$ estimates the optimal cost-to-come from the initial state $s_0$ to state $s$:
\begin{equation} 
    \label{eq:g_value}
    g(s) = \min_{\tau: s_0 \rightsquigarrow s} \sum_{t=0}^{|\tau|-1} c(s_t, a_t), \nonumber
\end{equation}
and $h(s)$ is a heuristic estimating the cost-to-go from $s$ to any goal state:
\begin{equation}    
    \label{eq:heuristic-func}
    h(s) \leq \min_{s_g \in \mathcal{G}} \min_{\tau: s \rightsquigarrow s_g} \sum_{t=0}^{|\tau|-1} c(s_t, a_t).
\end{equation}

\noindent At each iteration, A* selects the state with minimum $f$-value for expansion, generating successor states through available actions and updating their cost-to-come estimates. This process continues until a goal state is reached, guaranteeing optimal solutions when the heuristic is admissible.

\textbf{Connection to Reinforcement Learning:}  Our approach is based on the mathematical equivalence between search heuristics and RL value functions. Specifically, the heuristic function $h(s)$ in Equation~\eqref{eq:heuristic-func} estimates a quantity similar to the state value function of RL when the rewards are formulated as negative values.
\begin{equation}
    V^{\pi}(s) = \mathbb{E}_{\tau \sim \pi}\left[\sum_{t=0}^{\infty} \gamma^t R(s_t, a_t) \mid s_0 = s\right], \nonumber
\end{equation}
when we set $c(s,a) = -R(s,a)$ (converting rewards to costs). This equivalence allows us to use the critic network $Q_{\phi}(s,a)$ as a learned heuristic, while the actor network $\pi_{\theta}(a|s)$ can guide action selection during state expansion, replacing hand-crafted action spaces with learned action distributions.

\subsection{ePA*SE}
One particular important algorithm that motivated this work is ePA*SE~\cite{mukherjee2022epa}: a parallel version of edge-A* that addresses expensive edge evaluation through two key innovations. First, it decouples state expansion from edge evaluation by maintaining an open list of edges instead of states, using edge placeholders to separate successor generation from edge evaluation. Second, it enables parallelization of edge evaluations where each computational thread evaluates a single edge at a time, distributing the computationally expensive workload across threads for greater efficiency. The algorithm selects the edge $e = (s, a)$ for expansion with the smallest priority value $f(e) = g(s) + w \cdot h(s)$, where $g(s)$ is the cost-to-come, $h(s)$ is the heuristic estimate of the source state $\state$, and $w$ is the heuristic weight, which controls the optimality bound and the greediness of the search. During the expansion of edge $e$, the algorithm evaluates the edge to determine the resulting successor state $s'$, then generates all outgoing edges from $s'$ and inserts them into the open list.

\section{Algorithmic Approach} \label{Sec:approach}


\begin{figure}[!t]
\vspace{1em}
    \centering
    \includegraphics[width=\columnwidth]{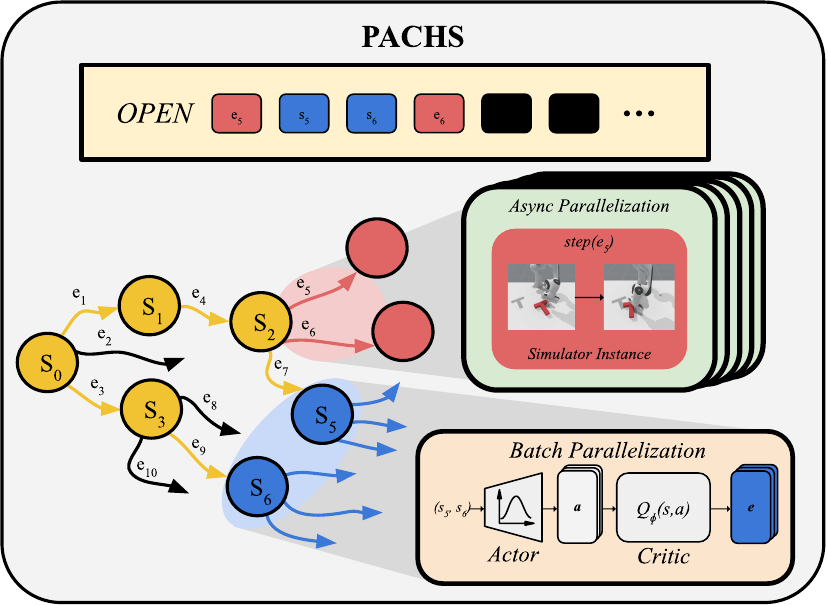}
    \caption{\pachs{} multi-level parallelization. Yellow elements represent completed expansions and evaluations. The \open list maintains edge candidates ordered by a priority function. Red edges ($\edge_5$, $\edge_6$) undergo parallel evaluation across threads, while blue states ($\state_5$, $\state_6$) have their actions and heuristics generated in batches. Both processes execute simultaneously, demonstrating CPU thread-level and GPU batch-level parallelization.
    }
    \label{pachs/fig/pachs_expand}
\vspace{-2em}
\end{figure}

To address the limitations of RL deployment and traditional best-first search, we present \pachs{}, a best-first heuristic search framework that enables multi-step reasoning during RL inference. Our approach capitalizes on the natural correspondence between RL and search concepts: the actor network generates candidate actions (edges) during state expansion, while the critic network provides learned heuristics to guide the search process. This integration leverages the knowledge acquired during RL training while extending its capabilities and performance during inference.

Enabling effective search in robotic domains requires addressing significant computational challenges. Neural network inference for both actor and critic models is computationally expensive, and robotic tasks often involve costly edge evaluations through physics simulation, collision checking, and dynamics computation. Without sophisticated parallelization strategies, search-based approaches can be prohibitively slow for real-time deployment. Therefore, \pachs{} incorporates multi-level parallelization to make search with learned modules practical for robotics (Fig. \ref{pachs/fig/pachs_expand}).

Inspired by ePA*SE, \pachs{} decouples edge evaluation from state expansion to allow for a finer grain of parallelization. \pachs{} uses the actor to generate actions in continuous space and a critic to evaluate the priority of generated actions. Since the actor and critic are trained with a physics simulator, they implicitly learn and embed the dynamics of the environment, waiving the engineering effort to design handcrafted heuristic functions and define discrete action spaces, resulting in learned, dynamically-generated motion primitives.

\begin{algorithm}[b!]
\label{alg:pachs-plan}
\begin{footnotesize}
\caption{PACHS: Planning Loop}
\begin{algorithmic}[1]
\State \textbf{INPUT:}
\State $\qquad \pi$: Stochastic Actor Policy
\State $\qquad \mathcal{Q}_{\phi}$: Q-value Critic Network
\State $\qquad \mathbf{s}_0$: start state
\State $\qquad \mathcal{G}$: goal condition
\State $\qquad N_t$: number of thread
\State \textbf{OUTPUT:}
\State $\qquad \mathcal{P}$: Path from start to goal

\vspace{4pt}

\Procedure{Plan}{}
    \State \textit{OPEN} = $\emptyset$
    \State $terminate$ = False
\State insert $(\mathbf{s}_0, \mathbf{a}^\mathbf{d})$ in \textit{OPEN} \Comment{{\color{orange} \scriptsize Dummy edge from $\mathbf{s}_0$}} \label{alg:pachs-plan/dummy}
    \State LOCK
    \While{\textbf{not} $terminate$}
        \If{\textit{OPEN} $= \emptyset$}
            \If{\Call{NoWIP}{}} \Comment{{\color{orange} \scriptsize None of the threads evaluate edges}} \label{alg:pachs-plan/checkwork}
                \State $terminate = $ True
                \State UNLOCK
                \State \Return $\emptyset$ \Comment{{\color{orange} \scriptsize Exhausted open list, no solution found}}
            \Else
                \State UNLOCK
                \State wait until \textit{OPEN} change \Comment{{\color{orange} \scriptsize Open list is empty, wait until new edge is added to the open list}}
            \EndIf
        \EndIf
        \State pop the min edge $(\mathbf{s}, \mathbf{a})$ from \open \label{alg:pachs-plan/pop-min}
        \If{$\mathcal{G}(\mathbf{s}) = $ True} \label{alg:pachs-plan/check-goal} \Comment{{\color{orange} \scriptsize $\textbf{s}$ satisfy the goal condition}}
            \State $terminate =$ True 
            \State UNLOCK
            \State \Return $\mathcal{P}$ = \textsc{Backtrack}($\mathbf{s}$) \Comment{{\color{orange} \scriptsize Reconstruct the path}}
        \EndIf
        \State UNLOCK
        \While{$(\mathbf{s}, \mathbf{a})$ has not been assigned a thread}
            \For{$i = 1 : N_t$}
                \If{thread $i$ is available}
                    \If{thread $i$ has not been spawned}
                        \State Spawn \textsc{EdgeExpandThread}($i$)
                    \EndIf
                    \State Assign $(\mathbf{s}, \mathbf{a})$ to thread $i$ \Comment{{\color{orange} \scriptsize Asynchronous operation}}
                \EndIf
            \EndFor
        \EndWhile
        \State LOCK
    \EndWhile
    \State UNLOCK
\EndProcedure
\vspace{4pt}
\Procedure{NoWIP}{}
    \For{t in SpawnedThreads}
        \If{t is Working}
        \State \Return False
        \EndIf
    \EndFor
    \State \Return True
\EndProcedure
\end{algorithmic}
\end{footnotesize}
\end{algorithm}

\begin{algorithm}[t]
\label{alg:pachs-expand}
\begin{footnotesize}
\caption{PACHS: Expansion}
\begin{algorithmic}[1]
\vspace{4pt}
\Procedure{ExpandEdgeThread}{$i$}
    \While{\textbf{not} $terminate$}
        \If{thread $i$ has been assigned an edge $(\mathbf{s}, \mathbf{a})$}
            \If{$\mathbf{a} = \mathbf{a^d}$}
                \State \Call{ExpandState}{$\mathbf{s}$}
            \Else
                \State \Call{ExpandEdge}{$(\mathbf{s}, \mathbf{a})$}
            \EndIf
        \EndIf
    \EndWhile
\EndProcedure

\vspace{4pt}

\Procedure{ExpandState}{$\mathbf{s}$}
    \State $\mathbf{\Vec{a}}$ = $\pi(\mathbf{s})$ \Comment{{\color{orange} \scriptsize Sample a batch of actions}} \label{alg:pachs-expand/action-gen}
    \State $\mathbf{\Vec{q}}$ = $\mathcal{Q^{\phi}}(\mathbf{s}, \mathbf{\Vec{a}})$ \Comment{{\color{orange} \scriptsize Compute the Q-value in a batch}} \label{alg:pachs-expand/q-compute}
    \State LOCK \Comment{{\color{orange} \scriptsize Expensive neural network operations are non-blocking}}
    \For{$\mathbf{a_i}$ in $\mathbf{\Vec{a}}$}
        \State $f((\mathbf{s}, \mathbf{a_i})) = g(\mathbf{s}) + w*\mathbf{q_i}$ \Comment{{\color{orange}  \scriptsize Each edge has a different priority based on the estimation from critic}} \label{alg:pachs-expand/edge-prior}
        \State insert $(\mathbf{s}, \mathbf{a_i})$ in \textit{OPEN} with $f((\mathbf{s}, \mathbf{a_i}))$
    \EndFor
    \State insert $\mathbf{s}$ in \textit{CLOSED} \label{alg:pachs-expand/mark-closed}
    \State UNLOCK
\EndProcedure

\vspace{4pt}

\Procedure{ExpandEdge}{$(\mathbf{s}, \mathbf{a})$}
    \State $\mathbf{s}', c((\mathbf{s}, \mathbf{a})) = $ \Call{Evaluate}{$(\mathbf{s}, \mathbf{a})$} \label{alg:pachs-exp/evaluate}
    \State LOCK
    \If{$\mathbf{s}' \notin \textit{CLOSED}$ and $g(\mathbf{s}') > g(\mathbf{s}) + c((\mathbf{s}, \mathbf{a}))$}
        \State $g(\mathbf{s}') = g(\mathbf{s}) + c((\mathbf{s}, \mathbf{a}))$
        \State $\mathbf{s}'.\text{parent} = \mathbf{s}.\text{id}$ \label{alg:pachs-expand/mark-parent}
        \State $f((\mathbf{s}', \mathbf{a}^\mathbf{d})) = g(\mathbf{s}') + w*(\textbf{q} - c((\mathbf{s}, \mathbf{a})))$ \Comment{{\color{orange} \scriptsize Reusing Q-value of the edge to calculate heuristic for the successor state}}
        \State insert/update $(\mathbf{s}', \mathbf{a}^\mathbf{d})$ in \textit{OPEN} with $f((\mathbf{s}', \mathbf{a}^\mathbf{d}))$ \Comment{{\color{orange} \scriptsize The dummy action indicates the edge is a representation of a state}} \label{alg:pachs-expand/state-prior}
    \EndIf
    \State UNLOCK
\EndProcedure
\end{algorithmic}
\end{footnotesize}
\end{algorithm}

\pachs{} addresses a key limitation in edge prioritization that arises when edge costs are unavailable until full evaluation. In domains using physics simulation to evaluate action outcomes, ePA*SE does not evaluate edges before selection from the \open list, and its edge priorities are of the form $f(e) = f((s, a)) = g(s) + w \cdot h(s)$, which lack action-specific information. This causes all outgoing edges from the same parent state to have identical priorities, making edge selection depend solely on tie-breaking strategies.
\pachs{} overcomes this limitation by incorporating learned Q-values (the critic outputs) into the priority function. The edge priority in \pachs{} is $f(e) = g(s) + w \cdot  Q_{\phi}(s, a) \approx g(s) + w \cdot (c((s, a)) + h(s'|a))$. This formulation leverages the critic's learned cost estimation to prioritize promising actions during search, enabling more informed edge selection while maintaining the computational benefits of lazy state and edge generation.

Algorithms 1 \& 2 detail \pachs{}'s operation. Algorithm 1 handles the main planning loop on the primary thread, while Algorithm 2 manages parallel edge expansion across worker threads.

\textbf{Main Planning Loop (Alg. 1):} The algorithm maintains an \open list of edges prioritized by their $f$-values. States are represented as edges by pairing them with dummy actions $\mathbf{a^d}$ (Line \ref{alg:pachs-plan/dummy}). A dummy edge $(\mathbf{s}, \mathbf{a^d})$ serves as a placeholder that indicates state $\mathbf{s}$ is ready for expansion--when selected, it triggers the generation of real outgoing actions from that state rather than executing a specific action. This design enables unified treatment of states and actions within the edge-based search framework. The main loop selects the highest priority edge (Line \ref{alg:pachs-plan/pop-min}), checks for goal satisfaction (Line \ref{alg:pachs-plan/check-goal}), and assigns edges to available worker threads for parallel processing. When a goal is reached, the solution path is reconstructed by backtracking parent pointers from the goal state to the start state.

\textbf{Worker Thread Operations (Alg. 2):} Worker threads handle two types of edges. \textbf{Dummy edges} trigger state expansion: the actor generates action batches (Line \ref{alg:pachs-expand/action-gen}) from a continuous action distribution, the critic computes the corresponding Q-values (Line \ref{alg:pachs-expand/q-compute}), and real edges are inserted into the \open list with priorities $f((\mathbf{s}, \mathbf{a_i})) = g(\mathbf{s}) + w \cdot \mathbf{q_i}$ (Line \ref{alg:pachs-expand/edge-prior}). The state is then marked as \closed (Line \ref{alg:pachs-expand/mark-closed}). \textbf{Real edges} undergo evaluation (Line \ref{alg:pachs-exp/evaluate}) to compute successor states and edge costs through physics simulation. When improvements are found, the algorithm updates the successor's $g$-value, parent pointer (Line \ref{alg:pachs-expand/mark-parent}), and inserts a dummy edge for the successor with updated priority (Line \ref{alg:pachs-expand/state-prior}).

\textbf{Parallelization:} Critical data structure updates use locks to prevent race conditions, while expensive operations (neural network inference and physics simulation) run lock-free for maximum parallel efficiency.

\subsection{Real-time Adaptation} \label{Sec:approach/real-time}
To deploy \pachs{} in real-world scenarios involving robot–object interactions, the system must operate in a real-time, closed-loop manner. We therefore implement a real-time version of \pachs{}. Given a time budget for a single planning query (e.g., 3 seconds), the robot observes the current environment state, plans within the allotted time, and reconstructs the best path it has found. The algorithm may not always find a complete path to the goal within this limit. If the budget expires before a path to the goal is found, the algorithm selects the edge with the best f-value from the open list, reconstructs a partial path from the current state to this edge, and executes the first few actions of that path. This process repeats--plan, execute, observe--until the goal is reached.
\section{Experiments}

\begin{figure}[t]
\vspace{1em}
    \centering
    \includegraphics[width=\columnwidth]{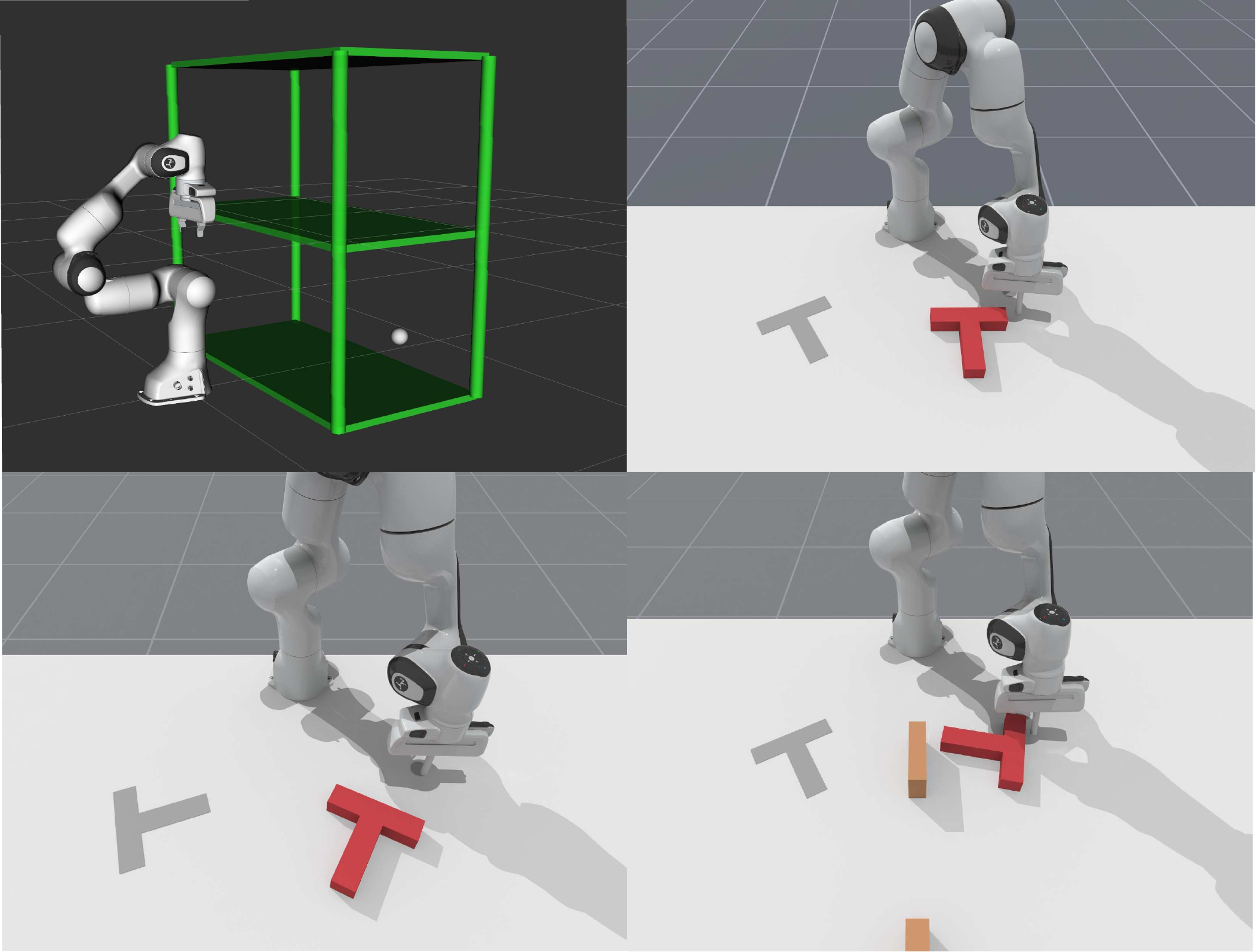}
    \caption{Our four simulated environments. Upper left: \tps---find a collision-free motion plan from the start state to a random EE target position within the shelves (white marker). Upper right: \ttf---Push the T-shaped object from a random start state to a fixed target pose (gray T shape). Bottom left: \ttr---push T from a random start pose to a random goal pose. Bottom right: \tto---push T from a random start pose to a fixed target pose with added obstacles--the T-shaped object must be pushed between the blocks to reach the goal.}
    \label{fig:panda-tasks}
    \vspace{-1em}
\end{figure}

We evaluated \pachs{} using a 7-DoF Franka Panda arm in four different tasks across two domains as shown in Fig. \ref{fig:panda-tasks}. The first domain, \tps, is collision-free motion planning in a shelf environment, where the planner must find a feasible path from a home configuration to a desired end-effector (EE) position within the shelf's reachable workspace. The second domain involves Push T tasks with three variants: \ttf, \ttr, and \tto. All tasks begin with the T-shaped object in a random initial pose. In \ttf and \tto, the robot must push the object to a fixed target pose, while \ttr requires pushing to a random goal pose. The \tto variant additionally includes static obstacles in the environment.
We trained SAC models for the \tps, \ttf, and \ttr tasks. To evaluate generalization capabilities, we applied the \ttf models to the \tto task without retraining. Training details and model specifications are provided in Sections \ref{Sec:Task-Shelf} and \ref{Sec:Task-PushT}. All actors are trained to output a 7-DoF delta joint position control for the Panda arm.
All models were implemented in PyTorch \cite{pytorch}, while search algorithms were implemented in C++. Experiments were conducted on a system equipped with an Intel i7-11800H CPU and NVIDIA GeForce RTX 3070 laptop GPU.

\subsection{Baseline Methods}
We compare \pachs{} against four baseline algorithms to assess deployment efficiency and its ability to generalize in inference, without retraining. \textbf{\rollout} represents the standard sequential rollout approach that maintains a single environment and executes the Observe-GetAction-StepEnv cycle until finding a solution or reaching the maximum step limit, which we set equal to the training episode length. \textbf{\prollout} is a parallel rollout method that instantiates multiple environments with identical problem instances and performs batch rollouts. We enforce a total evaluation step budget to ensure fair computational comparisons. When a rollout batch reaches the step limit without finding a solution, episodes restart and continue until either a solution is found or the budget is exhausted. \textbf{\pbeam} implements parallel beam search that uses the actor to sample actions, then selects a beam width of states based on Q-values estimated by the critic. We also compare against \textbf{\epase} specifically for the \tps environment, where a suitable heuristic function is readily available without extensive engineering effort. In this case, we employ a 3D breadth-first search algorithm to compute the voxel-based distance from the current EE state to the goal state.

\subsection{Panda Shelf Motion Planning} \label{Sec:Task-Shelf}
The main purpose of collision-free motion planning experiment is to test whether \pachs{}, that utilizes neural networks during search, is efficient enough compared to existing methods.
We developed the \tps environment based on the panda-gym package \cite{pandagym} and trained the SAC model using the stable-baseline3 framework \cite{stable-baselines3}. We used two fully connected layers with 64 neurons for both the actor and the critic. We used a dense reward that is a linear combination of the negated Euclidean distance from the robot's current EE position to the desired end-effector position, and a penalty term that is the minimum distance between the bookshelf and the arm, to encourage motion that avoids collision. We trained the models with 1 million training steps.

\begin{figure}[t]
\vspace{1em}
    \centering
    \includegraphics[width=\columnwidth]{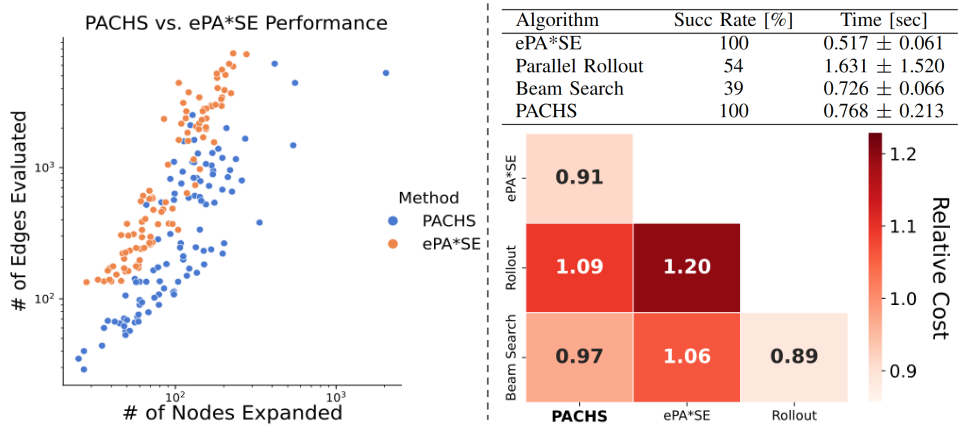}
    \caption{Results for collision-free motion planning in the Panda shelf environment.
    Upper right: Success rate and planning time for each algorithm. Only eP*ASE and \pachs{} consistently find successful motions. Unlike ePA*SE, however, \pachs{} uses two neural networks during planning—one for action generation and one for cost-to-go evaluation—computationally expensive processes. Despite this overhead, its planning time remains comparable. Lower right: Confusion matrix (row to column ratio) showing that \pachs{} produces solutions with costs similar to the baselines. Left: \pachs{} uses learned modules that focus the search, resulting in fewer evaluated nodes per number of expanded nodes, supporting the use of the critic network to estimate edge costs and prioritize them for evaluation.}
    \label{fig:shlef_result}
    \vspace{-1em}
\end{figure}

We generate 100 problem instances of an EE target position within the shelf which is reachable by the robot. All planners are given a 60 seconds to solve.
Since the collision-free motion planning does not require simulating actions, we use a collision checker to evaluate edges. In the case of parallel rollout, the rollout instantly restarts when it leads to a collision. Results are shown in Fig. \ref{fig:shlef_result}. The cost is the accumulated distance of the end-effector throughout the motion. The cost and time only count the instances for which the associated planner found a solution.

We observe that ePA*SE and \pachs{} achieve 100\% success rates, while parallel rollout and beam search show considerably lower performance. Despite the computational overhead from using neural networks, \pachs{} matches ePA*SE's solution quality and planning time. This efficiency comes from the critic network's guidance. For identical numbers of expanded nodes, \pachs{} evaluates substantially fewer edges, illustrating how its prioritization reduces computational waste.

\begin{figure*}[t]
\vspace{1em}
    \centering
    \includegraphics[width=0.325\textwidth]{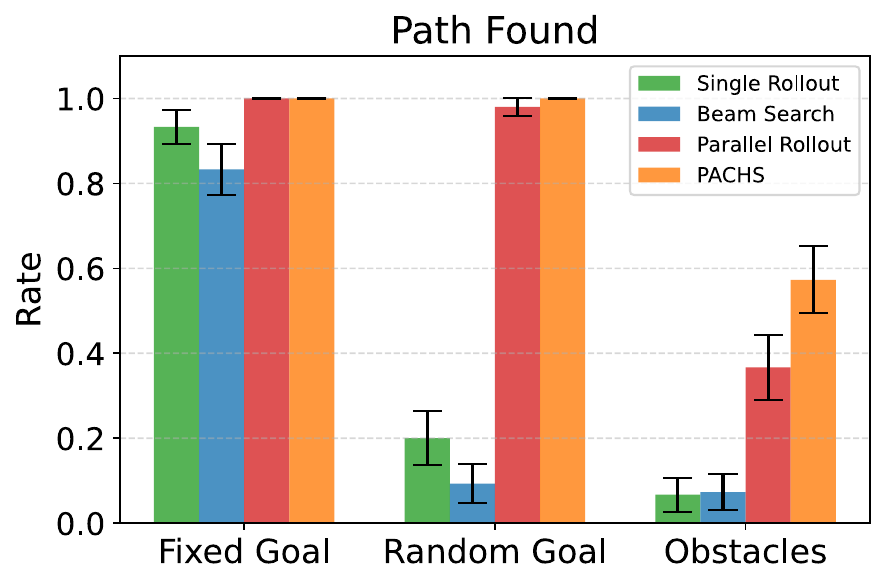}
    \includegraphics[width=0.325\textwidth]{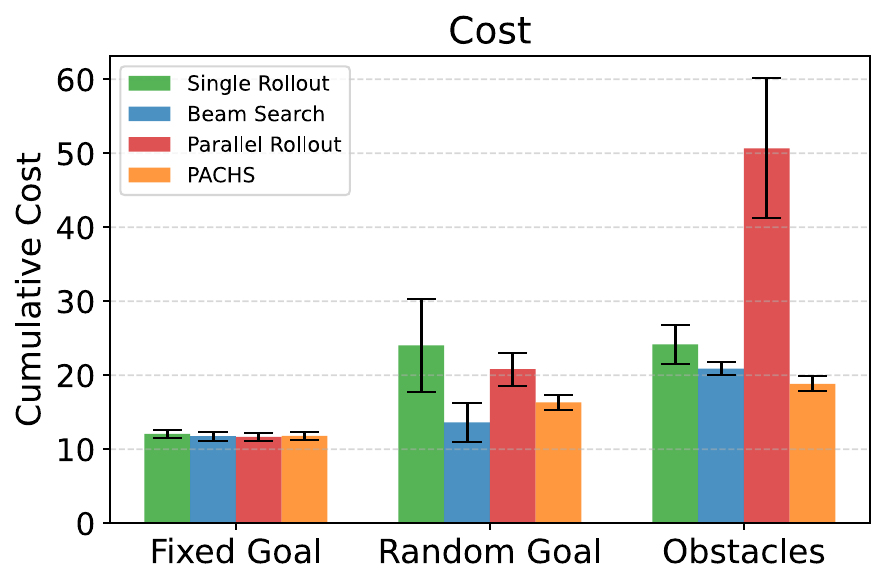}
    \includegraphics[width=0.325\textwidth]{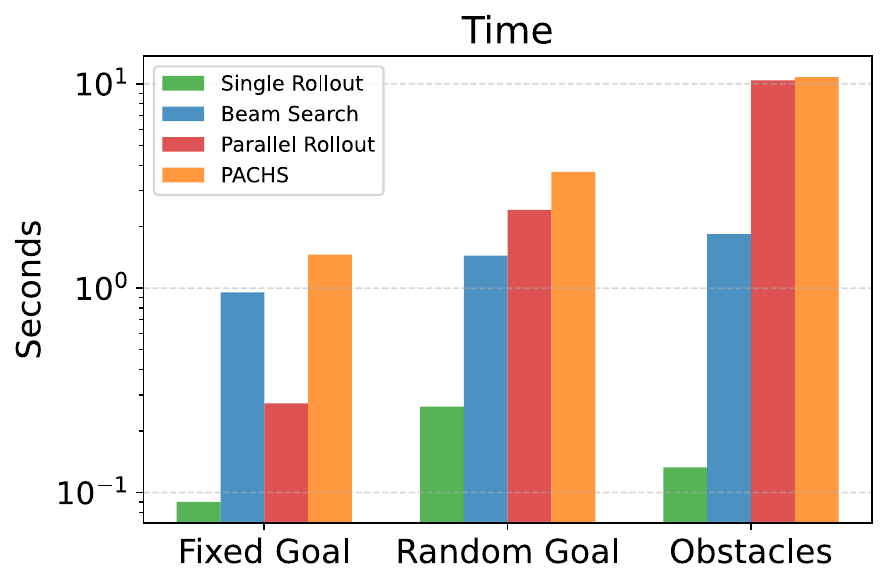}
    \caption{Algorithms comparison across three push-T task domains: \ttf, \ttr, \tto. We test each planner's ability to find a plan within a budget of 100,000 edge evaluations, without considering execution. For each scenario, we generate 30 different instances and run each 5 times to obtain average and confidence interval statistics, totaling 150 runs per scenario and planner. As shown in the left plot, while policy rollout achieves 93\% success rate in the fixed goal case, \pachs{} achieves 100\%. In the random goal case, despite the policy's 20\% success rate, \pachs{} achieves 100\%. When adding obstacles to the fixed goal environment, \pachs{} generalizes significantly better than both single rollout and parallel rollout methods. The planning time and cost results shown are based only on successful runs, and while PACHS has slower planning times, it achieves substantially lower costs and solves more instances overall.}
    \label{pachs/fig/pust_open_time}
    \vspace{-1em}
\end{figure*}

\subsection{Push T: Solution Finding Evaluation} \label{Sec:Task-PushT}
The objective of this experiment is to assess the ability of \pachs{}, compared to baseline approaches, to find solutions to manipulation tasks that require interactions with the environment, and its ability to generalize during inference without network finetuning.
The PushT environment tasks are based on the PushT task of ManiSkill~\cite{mu2021maniskill}. The 7-DoF Panda arm must manipulate a T-shaped object by moving it from an initial configuration to a target configuration on a planar surface (gray projection as shown in Fig. \ref{fig:panda-tasks}). The state of the object is defined by the position of its center of mass and its orientation. Task success is achieved when at least 90\% of the goal configuration’s shadow projection on the table is covered by the object.

For the easier problem, \ttf, we trained a 3-layer fully-connected network for both the actor and the critic, with 256 neurons in each layer and a ReLU activation function. For the harder \ttr environment, each layer size is 1024. The SAC model was trained with a negative dense reward formulation composed of three terms: the angular distance between the T-object projected on the table and its goal state, the distance function between the T-object and its goal state, and the distance between the end effector and the T-object (a commonly used reward function for the Push-T environment). 
For all tasks in the Push T domain, we randomly sampled 30 problem instances. These correspond to 30 random initial T poses in \ttf and \tto and 30 pairs of random initial and goal T poses in \ttr. The positions of the obstacles in \tto remain the same across different problems. We ran the 30 problem instances 5 times to obtain statistics for the success rate of the algorithm returning a solution. We gave \pachs{} and parallel rollout a total evaluation budget of 100k. The experimental results are shown in Fig. \ref{pachs/fig/pust_open_time}. We observe that \pachs{} consistently solves more problem instances (Path Found metric). Interestingly, despite the theoretical advantages of best-first search methods, Beam Search shows the lowest performances across all planners. This highlights that not all search strategies are beneficial for solving such tasks.
Notably, \pachs{} achieves a significantly higher success rate in \tto task, demonstrating the generalization capabilities it can provide to RL models, allowing them to solve tasks beyond their training domain.


\begin{wrapfigure}{l}{0.6\linewidth}
  \begin{center}
    \vspace{-14pt}
    \hspace{-8pt}
    \includegraphics[width=\linewidth]{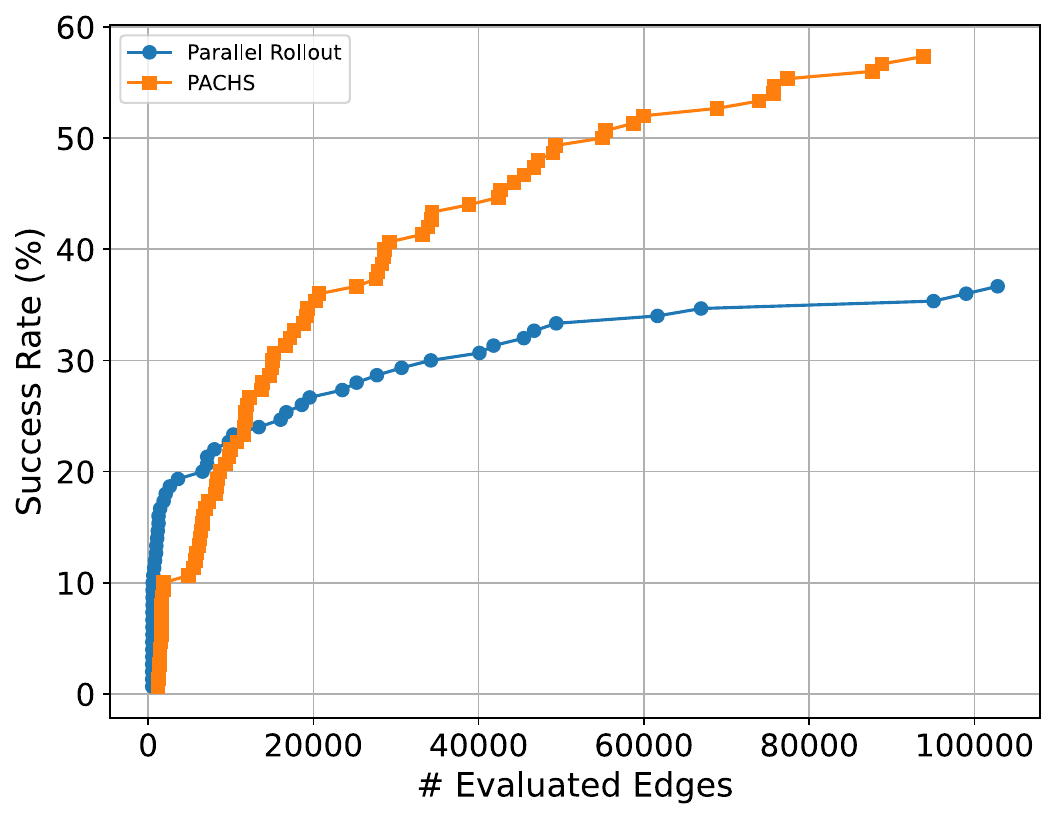}
    \hspace{-8pt}
    \vspace{-12pt}
    \label{pachs/fig/pust_open_evolution}
  \end{center}
\end{wrapfigure}
\noindent The left figure plots success rate versus number of edges evaluated for \pachs{} and parallel rollout in the \tto task. Each dot represents the percentage of instances solved after x edge evaluations. We observe that parallel rollout's success rate spikes quickly at lower evaluation budgets but plateaus as the edge evaluation budget increases. In contrast, \pachs{} requires slightly more evaluations to match parallel rollout's early performance, yet demonstrates steady improvement as the evaluation budget grows. This reveals an important insight: randomness provides only limited generalization during deployment, highlighting the advantages of structured exploration through search.


\subsection{Push T: Execution Performance Evaluation}

This experiment investigates whether \pachs{}'s planning improvements translate to better closed-loop execution performance--a critical requirement for real-world robotic deployment. Building on Section \ref{Sec:Task-PushT} showing that \pachs{} finds solutions and improves generalization for RL deployment, we now evaluate whether these benefits extend to dynamic replanning scenarios where the robot must continuously adapt to environmental feedback.
In this experiment, we maintain an independent simulator environment to mimic a real-world scenario, which we call the ``world'' simulator. Then, each algorithm is given a short planning budget to plan a full path or a partial path. After a path is returned, a certain horizon of actions from the plan are executed in the ``world" simulator, which then provides the observation for the next planning query. We compare single rollout in the ``world" simulator, parallel rollout, and \pachs{}. If no plan is found within the time budget, \pachs{} will return a partial path to the best state, described in Sec. \ref{Sec:approach/real-time}. Similarly, parallel rollout will keep track of the rollout that has the best cumulated reward and will return that path if no rollout leads to a solution.

Similar to Sec. \ref{Sec:Task-PushT}, we run five rounds of evaluation with 20 problem instances. \pachs{} and parallel rollout are given 3 seconds of planning budget and an action execution horizon of 10 (i.e., after each planning query, 10 actions are executed in the ``world" simulator). We limit the process to 30 replanning calls, with successful execution determined by reaching the goal in the ``world'' simulator.

\begin{figure}[t]
\vspace{1em}
    \centering
    \includegraphics[width=0.8\columnwidth]{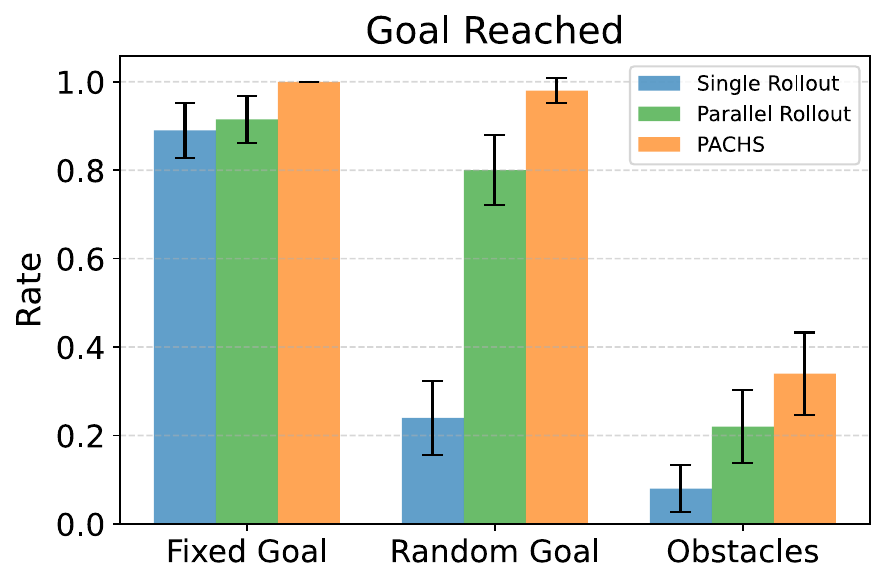}
    \caption{Comparison of the closed-loop success rate of single rollout, parallel rollout, and \pachs{} for all Push T tasks. \pachs{} is consistently better than baselines across different tasks.}
    \label{pachs/fig/pust_close_cost}
    \vspace{-1em}
\end{figure}

Fig. \ref{pachs/fig/pust_close_cost} shows the success rate of reaching the goal in the ``world" simulator over the three Push T tasks. The performance of \pachs{} is consistently superior compared to both rollout approaches, demonstrating a better ``sim-to-sim" transfer ability. Interestingly, even though the parallel rollout achieved an impressive success rate at finding the solution in Fig. \ref{pachs/fig/pust_open_evolution}, its performance drops significantly while \pachs{} maintains similar performance when executed in a closed-loop manner, indicating that the solutions found by rollout are inconsistent compared to \pachs{}.



\section{Conclusion}

This work presents \pachs{}, a novel framework that integrates actor-critic reinforcement learning models with parallel best-first search to enable zero-shot generalization during inference.
By leveraging the actor network for action generation and the critic network as a learned heuristic function, \pachs{} addresses fundamental limitations in RL deployment while requiring no model retraining.
\pachs{}'s parallelization strategy incorporates both CPU thread-level and GPU batch-level optimization, resulting in efficient search-based inference that is practical for deployment.
Experimental evaluation across collision-free motion planning and contact-rich manipulation demonstrates improvement in success rates and generalization, particularly in complex environments with obstacles not encountered during training.
In the future, we plan to deploy \pachs{} to real robots to evaluate generalization capabilities for sim2real transfer.
We also plan to investigate additional heuristics or provable guarantees on the search.

\bibliographystyle{ieeetr}
\bibliography{icra_paper}

\end{document}